\pdfoutput=1

\documentclass[11pt]{article}

\usepackage{naacl2021}

\usepackage{times}
\usepackage{latexsym}
\usepackage{graphicx}
\usepackage{booktabs}
\usepackage[ruled,vlined]{algorithm2e}

\usepackage[T1]{fontenc}

\usepackage[utf8]{inputenc}

\usepackage{microtype}
\usepackage{tikz-dependency}
\usepackage[linguistics]{forest}
\usepackage{caption}
\usepackage{subcaption}

\newif\ifdraft
\drafttrue
\definecolor{dkgreen}{RGB}{0,130,0}

\definecolor{mygreen}{HTML}{008000}

\newcommand{\interalia}[1]{\citep[\emph{inter alia}]{#1}}
\newcommand{\sub}{\textsc{Sub}$^2$}

\newcommand{\noaug}{\textsc{NoAug}}
\newcommand{\synonym}{\textsc{CtxSub}}
\newcommand{\random}{\textsc{Rand}}

\newcommand{\randomword}{\textsc{RandWord}}

\title{Substructure Substitution: Structured Data Augmentation for NLP}

\author{Haoyue Shi \qquad Karen Livescu \qquad Kevin Gimpel \\ 
Toyota Technological Institute at Chicago, IL, USA, 60637 \\ 
{\tt\{freda,klivescu,kgimpel\}@ttic.edu}}

\begin{document}
\maketitle
\begin{abstract}
We study a family of data augmentation methods, substructure substitution (\sub), for natural language processing (NLP) tasks. 
\sub{} generates new examples by substituting substructures (e.g., subtrees or subsequences) with ones with the same label, which can be applied to many structured NLP tasks such as part-of-speech tagging and parsing. 
For more general tasks (e.g., text classification) which do not have explicitly annotated substructures, we present variations of \sub{} based on constituency parse trees, introducing structure-aware data augmentation methods to general NLP tasks. 
For most cases, training with the augmented dataset by \sub{} achieves better performance than training with the original training set. 
Further experiments show that \sub{} has more consistent performance than other investigated augmentation methods, across different tasks and sizes of the seed dataset. 
\end{abstract}

\begin{figure}[!ht]
    \small 
    \begin{subfigure}[b]{0.45\textwidth}
    \flushleft \emph{Original Sentences} \\
    \centering
    \begin{dependency}
    \begin{deptext}[column sep=0.1cm]
    I \& have \& \textcolor{blue}{a} \& \textcolor{blue}{book} \\
    PRP \& VBP \& \textcolor{blue}{DT} \& \textcolor{blue}{NN} \\
    \end{deptext} 
    \draw[blue,dashed] (0.1,0.4) rectangle(1.6,-0.4) node[pos=.5] {};
    \end{dependency}~~~
    \begin{dependency}
    \begin{deptext}[column sep=0.02cm]
    They \& ate \& \textcolor{orange}{an} \& \textcolor{orange}{orange} \\
    PRP \& VBD \& \textcolor{orange}{DT} \& \textcolor{orange}{NN} \\
    \end{deptext} 
    \draw[orange,dashed] (0,0.4) rectangle(1.65,-0.4) node[pos=.5] {};
    \end{dependency} \\
    \flushleft \emph{Generated Sentences} \\
    \centering
    \begin{dependency}
    \begin{deptext}[column sep=0.1cm]
    I \& have \& \textcolor{orange}{an} \& \textcolor{orange}{orange} \\
    PRP \& VBP \& \textcolor{orange}{DT} \& \textcolor{orange}{NN} \\
    \end{deptext} 
    \draw[orange,dashed] (0,0.4) rectangle(1.7,-0.4) node[pos=.5] {};
    \end{dependency}~~~
    \begin{dependency}
    \begin{deptext}[column sep=0.02cm]
    They \& ate \& \textcolor{blue}{a} \& \textcolor{blue}{book} \\
    PRP \& VBD \& \textcolor{blue}{DT} \& \textcolor{blue}{NN} \\
    \end{deptext} 
    \draw[blue,dashed] (0.15,0.4) rectangle(1.5,-0.4) node[pos=.5] {};
    \end{dependency} 
    \caption{Part-of-speech tagging.\label{fig:pos}}
    \end{subfigure}\\
    
    \begin{subfigure}[b]{0.4\textwidth}
    \flushleft \small \textit{Original sentences} \\
    \centering
    ~~~~~~~~~~\resizebox{0.45\textwidth}{0.045\textwidth}{
    \begin{forest}
     [ \large S
        [ \large NP
            [ {\large The}] [ {\large cat}]
        ]
        [ \large \textcolor{blue}{VP},tikz={\node [draw,blue,dashed,inner sep=0,fit to=tree]{};}
            [{\large \textcolor{blue}{is}}] [{\large \textcolor{blue}{sleeping}}]
        ]
     ]
    \end{forest}
    }~~~~~~
    \resizebox{0.4\textwidth}{0.045\textwidth}{
    \begin{forest}
     [ \large S
        [ \large NP
            [\large I]
        ]
        [ \large \textcolor{orange}{VP},tikz={\node [draw,orange,dashed,inner sep=0,fit to=tree]{};}
            [{\large \textcolor{orange}{love}}] 
            [{\large \textcolor{orange}{books}}]
        ]
     ]
    \end{forest}
    } \\
    \flushleft \small \textit{Original sentences} \\
    \centering
    ~~~~~~~~~~\resizebox{0.45\textwidth}{0.045\textwidth}{
    \begin{forest}
     [ \large S
        [ \large NP
            [ {\large The}] [ {\large cat}]
        ]
        [ \large \textcolor{orange}{VP},tikz={\node [draw,orange,dashed,inner sep=0,fit to=tree]{};}
            [{\large \textcolor{orange}{love}}] 
            [{\large \textcolor{orange}{books}}]
        ]
     ]
    \end{forest}
    }~~~~~~
    \resizebox{0.4\textwidth}{0.045\textwidth}{
    \begin{forest}
     [ \large S
        [ \large NP
            [\large I]
        ]
        [ \large \textcolor{blue}{VP},tikz={\node [draw,blue,dashed,inner sep=0,fit to=tree]{};}
            [{\large \textcolor{blue}{is}}] 
            [{\large \textcolor{blue}{sleeping}}]
        ]
     ]
    \end{forest}
    }
    \caption{Constituency parsing.\label{fig:c-parse}}
    \end{subfigure}\\
    
    \begin{subfigure}[b]{0.45\textwidth}
    \flushleft \emph{Original Sentences} \\
    \centering
    \begin{dependency}[arc edge, text only label, label style={above}]
    \begin{deptext}[column sep=0.2cm]
    My \& cat \& likes \& \textcolor{blue}{milk} \\
    \end{deptext} 
    \depedge{2}{1}{poss}
    \depedge{3}{2}{nsubj}
    \depedge[edge style=dashed,color=blue]{3}{4}{\textcolor{blue}{dobj}}
    \deproot[edge unit distance=0.2cm]{3}{root}
    \end{dependency}
    \begin{dependency}[arc edge, text only label, label style={above}]
    \begin{deptext}
    I \& read \& \textcolor{orange}{books} \\
    \end{deptext} 
    \depedge[edge style=dashed,color=orange]{2}{3}{\textcolor{orange}{dobj}}
    \depedge{2}{1}{nsubj}
    \deproot[edge unit distance=0.2cm]{2}{root}
    \end{dependency} \\
    \flushleft \emph{Generated Sentences}\\
    \centering
    \begin{dependency}[arc edge, text only label, label style={above}]
    \begin{deptext}[column sep=0.2cm]
    My \& cat \& likes \& \textcolor{orange}{books} \\
    \end{deptext} 
    \depedge{2}{1}{poss}
    \depedge{3}{2}{nsubj}
    \depedge[edge style=dashed,color=orange]{3}{4}{\textcolor{orange}{dobj}}
    \deproot[edge unit distance=0.2cm]{3}{root}
    \end{dependency}
    \begin{dependency}[arc edge, text only label, label style={above}]
    \begin{deptext}
    I \& read \& \textcolor{blue}{milk} \\
    \end{deptext} 
    \depedge[edge style=dashed,color=blue]{2}{3}{\textcolor{blue}{dobj}}
    \depedge{2}{1}{nsubj}
    \deproot[edge unit distance=0.2cm]{2}{root}
    \end{dependency} 
    \caption{Dependency parsing.\label{fig:d-parse}}
    \end{subfigure} \\
    
    \begin{subfigure}[b]{0.45\textwidth}
    \flushleft \emph{Original Sentences} \\
    \centering
    \begin{dependency}
    \begin{deptext}[column sep=0.1cm]
    I \& \textcolor{blue}{like} \& \textcolor{blue}{the} \& book \\
    \end{deptext} 
    \wordgroup[blue,dashed]{1}{2}{3}{a0}
    \end{dependency}~~~
    \begin{dependency}
    \begin{deptext}[column sep=0.02cm]
    I \& {like} \& \textcolor{orange}{the} \& \textcolor{orange}{movie} \\
    \end{deptext} 
    \wordgroup[orange,dashed]{1}{3}{4}{a0}
    \end{dependency} \\
    Label: positive ~~~~~~~ Label: positive
    \flushleft \emph{Generated Sentences} \\
    \centering
    \begin{dependency}
    \begin{deptext}[column sep=0.1cm]
    I \& \textcolor{orange}{the} \& \textcolor{orange}{movie} \& book \\
    \end{deptext} 
    \wordgroup[orange,dashed]{1}{2}{3}{a0}
    \end{dependency}~~~
    \begin{dependency}
    \begin{deptext}[column sep=0.1cm]
    I \& {like} \& \textcolor{blue}{like} \& \textcolor{blue}{the} \\
    \end{deptext} 
    \wordgroup[blue,dashed]{1}{2}{3}{a0}
    \end{dependency} \\
    Label: positive ~~~~~~~ Label: positive
    \caption{A variation of \sub{}: text sentiment classification with text spans as substructures, using the combination of (1) number of words and (2) text label as the substructure label.\label{fig:txt-class}}
    \end{subfigure}\\
    
    \caption{Illustration of \sub{} for investigated tasks. We generate new examples by same-label substructure substitution, whether or not the generated examples are semantically or syntactically acceptable. 
    }
    \label{fig:teaser}
\end{figure}
\section{Introduction}

Data augmentation has been shown effective for various natural language processing (NLP) tasks, such as machine translation \interalia{fadaee-etal-2017-data,gao-etal-2019-soft,xia-etal-2019-generalized}, text classification \citep{wei-zou-2019-eda,quteineh-etal-2020-textual}, semantic role labeling \citep{furstenau-lapata-2009-semi} and dialogue understanding \citep{hou-etal-2018-sequence,niu-bansal-2019-automatically}.
Such methods enhance the diversity of the training set by generating examples based on existing ones and simple heuristics, and make the training process more consistent \citep{xie2019unsupervised}. 
Most existing work focuses on word-level manipulation \interalia{kobayashi-2018-contextual,wei-zou-2019-eda,dai-adel-2020-analysis} or global sequence-to-sequence style generation \citep{sennrich-etal-2016-improving}.

In this work, we study a family of general data augmentation methods, substructure substitution (\sub), which generates new examples by same-label substructure substitution (Figure~\ref{fig:teaser}). 
\sub{} naturally fits structured prediction tasks such as part-of-speech tagging and parsing, where substructures exist in the annotations of the tasks. 
For more general NLP tasks such as text classification, we present a variation of \sub{} which (1) performs constituency parsing on existing examples, and (2) generates new examples by subtree substitution based on the parses.

Different from other investigated methods which sometimes hurt the performance of models, we show through intensive experiments that \sub{} helps models achieve competitive or better performance than training on the original dataset across tasks and original dataset sizes. 
When combined with pretrained language models \citep{conneau2019unsupervised}, \sub{} establishes new state of the art results for low-resource part-of-speech tagging and sentiment analysis. 

The question of whether explicit parse trees can help neural network--based approaches on downstream tasks has been raised 
by recent work \citep{shi-etal-2018-tree,havrylov-etal-2019-cooperative} in which non-linguistic balanced trees have been shown to rival the performance of those from syntactic parsers. 
Our work shows that constituency parse trees are more effective than balanced trees as backbones for \sub{} on text classification, especially when only few examples are available, introducing more potential applications for constituency parse trees in the neural network era. 

\section{Related Work}
Data augmentation aims to generate new examples based on available ones, without actually collecting new data.
Such methods reduce the cost of dataset collection, and usually boost the model performance on desired tasks. 
Most existing data augmentation methods for NLP tasks can be classified into the following categories: \\

\par\noindent\textbf{Token-level manipulation.}
Token-level manipulation has been widely studied in recent years. 
An intuitive way is to create new examples by substituting (word) tokens with ones with the same desired features, such as synonym substitution \cite{zhang2015character,wang-yang-2015-thats,fadaee-etal-2017-data,kobayashi-2018-contextual} or substitution with words having the same morphological features \citep{silfverberg-etal-2017-data}. 
Such methods have been applied to generate adversarial or negative examples which help improve the robustness of neural network--based NLP models \interalia{belinkov2018synthetic,shi-etal-2018-learning,alzantot-etal-2018-generating,zhang-etal-2019-generating-fluent,min-etal-2020-syntactic}, or to generate counterfactual examples which mitigate bias in natural language \citep{zmigrod-etal-2019-counterfactual,lu2020gender}. 

Other token-level manipulation methods introduce extra noise such as random token shuffling and deletion \citep{wang-etal-2018-switchout,wei-zou-2019-eda}. 
Models trained on the augmented dataset are expected to be more robust to the considered noise. \\

\par\noindent\textbf{Label-conditioned text generation.}
Recent work has explored generating new examples by training a conditional text generation model \interalia{bergmanis-etal-2017-training,liu-etal-2020-tell,ding-etal-2020-daga,liu-etal-2020-data}, or applying post-processing on the examples generated by pretrained models \citep{yang-etal-2020-generative,wan-etal-2020-improving,yoo-etal-2020-variational}. 
In the data augmentation stage, given labels in the original dataset as conditions, such models generate associated text accordingly.
The generated examples, together with the original datasets, are used to further train models for the primary tasks. 
A representative among them is back-translation \citep{sennrich-etal-2016-improving}, which has been demonstrated effective on not only machine translation, but also style-transfer \cite{prabhumoye-etal-2018-style,zhang-etal-2020-parallel}, conditional text generation \citep{sobrevilla-cabezudo-etal-2019-back}, and grammatical error correction \citep{xie-etal-2018-noising}.
 
Another group of work on example generation is to generate new examples based on predefined templates \cite{kafle-etal-2017-data,asai-hajishirzi-2020-logic}, where the templates are designed following heuristic, and usually task-specific, rules. \\
\par\noindent\textbf{Soft data augmentation.}
In addition to explicit generation of concrete examples, soft augmentation, which directly represents generated examples in a continuous vector space, has been proposed: \citet{gao-etal-2019-soft} propose to perform soft word substitution for machine translation; recent work has adapted the mix-up method \citep{zhang2018mixup}, which augments the original dataset by linearly interpolating the vector representations of text and labels, to text classification \citep{guo2019augmenting,sun-etal-2020-mixup}, named entity recognition \citep{chen-etal-2020-local} and compositional generalization \citep{guo-etal-2020-sequence}. \\
\par\noindent\textbf{Structure-aware data augmentation.}
Existing work has also sought potential gain from structures associated with natural language: \citet{xu-etal-2016-improved} improve word relation classification by dependency path--based augmentation. 
\citet{sahin-steedman-2018-data} show that subtree cropping and rotation based on dependency parse trees can help part-of-speech tagging for low-resource languages, while \citet{vania-etal-2019-systematic} has demonstrated that such methods also help dependency parsing when very limited training data is available. 

\sub{} also falls into this category. The idea of same-label substructure substitution has improved over baselines on structured prediction tasks such as semantic parsing \citep{jia-liang-2016-data}, constituency parsing \citep{shi-etal-2020-role}, dependency parsing \citep{dehouck-gomez-rodriguez-2020-data}, named entity recognition \citep{dai-adel-2020-analysis}, meaning representation--based text generation \citep{kedzie-mckeown-2020-controllable}, and compositional generalization \citep{andreas-2020-good}. 
To the best of our knowledge, however, \sub{} has not been systematically studied as a general data augmentation method for NLP tasks. 
In this work, we not only extend \sub{} to part-of-speech tagging and structured sentiment classification, but also present a variation that allows a broader range of NLP tasks (e.g., text classification) to benefit from syntactic parse trees. 
We evaluate \sub{} and several representative general data augmentation methods, which can be widely applied to various NLP tasks. 

When constituency parse trees are used, 
there is a connection between \sub{} and tree substitution grammars \citep[TSGs;][]{schabes1990mathematical}, where the approach can be viewed as (1) estimating a TSG using the given corpus and (2) drawing new sentences from the estimated TSG. 
\section{Method}
\label{sec:method}
We introduce the general framework we investigate in Section~\ref{sec:sub}, and describe the variations of \sub{} which can be extended to text classification and other NLP applications. 
\subsection{Substructure Substitution (\sub)}
\label{sec:sub}
As shown in Figure~\ref{fig:teaser}, given the original training set $\mathcal{D}$, \sub{} generates new examples using same-label substructure substitution, and repeats the process until the training set reaches the desired size. 
The general \sub{} procedure 
is presented in Algorithm~\ref{algo:sub}. 

\begin{algorithm}[ht]
\SetAlgoLined
\SetKwInOut{Input}{Output}
\KwIn{Original dataset $\mathcal{D}$, \\
desired dataset size $N > |\mathcal{D}|$}
\KwOut{Augmented dataset $\mathcal{D}'$}
 $\mathcal{D}' \leftarrow \mathcal{D}$; \\ 
 \Repeat{$|\mathcal{D}'| = N$}{
 Uniformly draw $s \in \textit{substructure}(\mathcal{D}')$ \\
 $S \leftarrow \textit{example}(s)$ \\
 Uniformly draw $u \in \{v \mid v \in \textit{substructure}(\mathcal{D}), \textit{label}(v) = \textit{label}(s), v\neq s\}$ \\
 $S' \leftarrow$ replace $s$ with $v$ in $S$ \\ 
 $\mathcal{D}' \leftarrow \mathcal{D}' \cup \{S'\}$ 
 }
 \caption{\label{algo:sub} \sub. }
\end{algorithm}

For part-of-speech (POS) tagging, we let text spans be substructures and use the corresponding POS tag sequence as substructure labels (Figure~\ref{fig:pos}); for constituency parsing, we use subtrees as the substructures, with phrase labels as the substructure labels (Figure~\ref{fig:c-parse}); for dependency parsing, we also use subtrees as substructures, and let the label of dependency arc, which links the head of the subtree to its parent, be the substructure labels.

\subsection{Variations of \sub{} for Text Classification}
\label{sec:var}
Text classification examples do not typically contain explicit substructures. However, 
we can obtain them by viewing all text spans as substructures (Figure~\ref{fig:txt-class}). This approach may be too unconstrained in practice and could introduce noise during augmentation, so 
we consider constraining substitution based on matching several features of the spans:
\begin{itemize}
    \item \textbf{Number of words}  (\sub+\textsc{n}): when considering this constraint, we can only 
    substitute a span with another having the same number of words; otherwise we can substitute a span with any other span. 
    \item \textbf{Phrase or not} (\sub+\textsc{p}): when considering this constraint, we can only substitute a phrase with another phrase (according to a constituency parse of the text); otherwise the considered spans do not necessarily need to be phrases. 
    \item \textbf{Phrase label} (\sub+\textsc{l}): this constraint is only applicable when also using \sub+\textsc{p}. When considering this constraint, we can only perform substitution between phrases with the same phrase label (from constituency parse trees). 
    
    \item \textbf{Text classification label} (\sub+\textsc{t}): when considering this constraint, we can only substitute a span with another span that comes from text annotated with the same class label as the original one; otherwise we can choose the alternative from any example text in the training corpus. 
\end{itemize}

We also investigate combinations of the above constraints, where we require all the involved substructures to be the same to perform \sub.
For example, \sub+\textsc{t+n} (Figure~\ref{fig:txt-class}) requires the original and the alternative span to have the same text label and the same number of words. 
\section{Experiments}
\label{sec:expr}
We evaluate \sub{} and other data augmentation baselines (Section~\ref{sec:expr-baseline}) on four tasks: part-of-speech tagging, dependency parsing, constituency parsing, and text classification. 

\subsection{Setup}
For part-of-speech tagging and text classification, we add a two-layer perceptron on top of XLM-R \citep{conneau2019unsupervised} embeddings, where we calculate contextualized token embeddings by a learnable weighted average across layers. 
We use endpoint concatenation (i.e., the concatenation of the first and last token representation) to obtain fixed-dimensional span or sentence features, and keep the pretrained model frozen during training.\footnote{We did not observe any significant improvement by finetuning the large pretrained language model, and for most cases, the performance is much worse than the current scheme we apply. }
For dependency parsing, we use the SuPar implementation of \citet{dozat2016deep}.\footnote{\url{https://github.com/yzhangcs/parser}} For constituency parsing, we use Benepar \citep{kitaev-klein-2018-constituency}.\footnote{\url{https://github.com/nikitakit/self-attentive-parser}}

For all data augmentation methods, including the baselines (Section~\ref{sec:expr-baseline}), we only augment the training set, and use the original development set. 
If not specified, we introduce 20 times more examples than the original training set when applying an augmentation method.
When introducing $k\times$ new examples, we also replicate the original training set  $k$ times to ensure that the model can access sufficient examples from the original distribution. 

All models are initialized with the XLM-R base model \citep{conneau2019unsupervised} if not specified. 
We train models for 20 epochs when applying the high-resource setting (i.e., high-resource part-of-speech tagging, sentiment classification trained on the full training set) or when applying data augmentation methods, and for 400 epochs in the low-resource settings without augmentation; we select the one with the highest accuracy or $F_1$ score on the development set. 
All models are optimized using Adam \citep{kingma2015adam}, where we try learning rates in $\{5\times 10^{-4}, 5\times10^{-5}\}$. 
For hidden size (i.e., the hidden size of the perceptron for part-of-speech tagging and text classification, the dimensionality of span representation and scoring multi-layer perceptron for constituency parsing, and the dimensionality of token representation and scoring multi-layer perceptron for dependency parsing), we vary between $128$ and $512$.  
We apply a 0.2 dropout ratio to the contextualized embeddings in the training stage. 
All other hyperparameters are the same as the default settings in the released codebases.

\subsection{Baselines}
\label{sec:expr-baseline}
We compare \sub{} to the following baselines:

\begin{itemize} 
\item \textbf{No augmentation} (\noaug), where the original training and development set are used. 
\item \textbf{Contextualized substitution} (\synonym), where we apply contextualized augmentation \citep{kobayashi-2018-contextual}, masking out a random word token from the existing dataset, and use multilingual-BERT  \citep[mBERT;][]{devlin-etal-2019-bert} to generate a different word. 
\item \textbf{Random shuffle} (\random), where we randomly shuffle all the words in the original sentence, while keeping the original structured or non-structured labels. 
It is worth noting that for dependency parsing, we shuffle the words, while maintaining the dependency arcs between individual words; for constituency parsing, we shuffle the terminal nodes, and insert them back into the tree structure. 
Our \random{} method for constituency parsing is arguably more noisy than that for dependency parsing. 
\end{itemize} 
For non-structured text classification tasks, we also introduce the following baselines: 

\begin{itemize} 
\item \textbf{Random word substitution} (\randomword), where we substitute a random word in an original example with another random word. 
This can be viewed as a less restricted version of \synonym. 

\item \textbf{Binary balanced tree--based \sub} (\sub+\textsc{p}, balanced tree). \citet{shi-etal-2018-tree} argue that binary balanced trees are better backbones for recursive neural networks \citep{zhu2015long,tai-etal-2015-improved} on text classification. In this work, we present binary balanced tree as the backbone for \sub{}: we (1) generate balanced trees by recursively splitting a span of $n$ words into two consecutive groups, which consist of $\left\lfloor\frac{n}{2}\right\rfloor$ and $\left\lceil\frac{n}{2}\right\rceil$ words respectively, and (2) treat each nonterminal in the balanced tree as a substructure to perform \sub. 
\end{itemize}

All of the data augmentation baselines are explicit augmentations where concrete new examples are generated and used. 
The methods above are generally applicable to a wide range of NLP tasks. 

\subsection{Part-of-Speech Tagging}
We conduct our experiments using the Universal Dependencies \citep[UD;][]{nivre-etal-2016-universal,nivre-etal-2020-universal}\footnote{\url{http://universaldependencies.org/}} dataset. 
\begin{table}[t]
    \centering \small
    \begin{tabular}{lcccc}
    \toprule
    Lang. & \multicolumn{1}{l}{\textsc{SotA}} & \multicolumn{1}{l}{mBERT} & \multicolumn{1}{l}{XLM-R} & \multicolumn{1}{l}{XLM-R} \\
    Aug. & \noaug & \noaug & \noaug & \sub \\
    \midrule
    \multicolumn{5}{l}{\emph{high-resource languages}}\\
    \midrule    
    avg. & 96.9 & 97.1 & \bf 97.7 & \bf 97.7 \\
    \midrule
    bg & 98.7 & 98.9 & \bf 99.4 & \bf 99.4 \\
    cs & 99.0 & 99.0 & \bf 99.2 & \bf 99.2 \\
    da & 97.2 & 97.8 & \bf 98.7 & 98.5 \\
    de & 94.4 & 94.6 & \bf 95.3 & 95.1 \\
    en & 96.1 & 96.5 & \bf 97.5 & 97.3 \\
    es & 96.8 & 96.9 & \bf 97.5 & \bf 97.5 \\
    eu & 96.1 & 95.7 & 96.6 & \bf 96.8 \\
    fa & 97.5 & 96.6 & \bf 98.6 & 98.5 \\
    fi & 95.8 & 96.9 & \bf 98.3 & \bf 98.3 \\
    fr & 96.6 & 96.7 & \bf 96.9 & \bf 96.9 \\
    he & 97.4 & 96.9 & \bf 97.9 & 97.8 \\
    hi & 97.4 & 96.9 & \bf 97.9 & 97.8 \\
    hr & 96.8 & 97.6 & 97.9 & \bf 98.0 \\
    id & 94.0 & 93.7 & \bf 93.8 & 93.7 \\
    it & 98.1 & 98.6 & \bf 98.7 & \bf 98.7 \\
    nl & 93.8 & 92.9 & \bf 94.0 & 93.6 \\
    no & 98.5 & 98.6 & \bf 99.0 & 98.9 \\
    pl & 97.7 & 98.5 & 98.8 & \bf 98.9 \\
    pt & 98.2 & 98.3 & \bf 98.6 & \bf 98.6 \\
    sl & 98.1 & 98.7 & \bf 99.2 & \bf 99.2 \\
    sv & 97.4 & 98.2 & \bf 98.9 & \bf 98.9 \\
    \midrule
    \multicolumn{5}{l}{\emph{low-resource languages}}\\
    \midrule
    avg. & 92.7 & 94.7 & 95.4 & \bf 96.1 \\
    \midrule
    el & 98.2 & 98.6 & \bf 98.8 & 98.7 \\
    et & 92.8 & 94.1 & 95.7 & \bf 96.3 \\
    ga & 91.1 & 92.9 & 94.1 & \bf 95.8 \\
    hu & 94.0 & 96.8 & \bf 97.7 & 97.5 \\
    ro & 91.5 & 95.0 & 94.9 & \bf 95.8 \\
    ta & 88.7 & 90.4 & 91.3 & \bf 92.5 \\
    \bottomrule
    \end{tabular}
    
    \caption[Table]{Part-of-speech tagging accuracy ($\times 100$) on the standard test set of UD 1.2 high-resource (top) and low-resource (bottom) languages, across different pretrained models and augmentation methods. The best numbers in each row are bolded. \textsc{SotA}: the best test accuracy for each language among all methods reported by \citet{heinzerling-strube-2019-sequence}. Note that XLM-R is the same setting as \noaug{} in Table~\ref{tab:pos-ud2.6}.}
    \label{tab:pos-ud1.2}
\end{table}

First, we compare both \noaug{} and \sub{} to the previous state-of-the-art performance \citep{heinzerling-strube-2019-sequence} to ensure that our baselines are strong enough (Table~\ref{tab:pos-ud1.2}). \citet{heinzerling-strube-2019-sequence} take the token-wise concatenation of mBERT last-layer representations, byte-pair encoding \citep[BPE;][]{gage1994new}--based LSTM hidden states and character-LSTM hidden states as the input to the classifier, and fine-tune the pretrained mBERT during training.
We find that with our framework with frozen mBERT and extra learnable layer weight parameters, we are able to obtain competitive or better results than those reported by \citet{heinzerling-strube-2019-sequence}; the gains grow larger when using XLM-R, which is trained on larger corpora than mBERT. 
In addition, by augmenting the training set with \sub{}, we obtain competitive performance on all languages, and achieve better average accuracy on low-resource languages. 

 \begin{table}[t]
 \small \centering 
\begin{tabular}{lcccc}
\toprule
Lang. & \multicolumn{1}{l}{\noaug} & \multicolumn{1}{l}{\synonym} & \multicolumn{1}{l}{\random} & \multicolumn{1}{l}{\sub} \\
\midrule 
avg. & 92.4 & 87.1 & 86.8 & \bf 93.0  \\
\midrule
be (hse) & 96.2 & 90.3 & 92.5 & \bf 96.9 \\
lt (hse) & 92.7 & 90.1 & 88.4 & \bf 93.1 \\
mr (ufal)& 87.9 & 81.5 & 84.5 & \bf 89.1 \\
ta (ttb) & 91.7 & 85.4 & 83.2 & \bf 92.3 \\
te (mtg) & \bf 93.8 & 88.2 & 85.6 & 93.0 \\ 
\bottomrule
\end{tabular}
 \caption{Part-of-speech tagging accuracy ($\times 100$) on the standard test set of selected UD 2.6 low-resource treebanks. The best number in each row is bolded. }
 \label{tab:pos-ud2.6}
\end{table}
We further test the part-of-speech tagging accuracy on 5 selected low-resource treebanks in the UD 2.6 dataset (Table~\ref{tab:pos-ud2.6}), following the official splits of the dataset. 
For four among the five investigated treebanks, \sub{} achieves the best performance among all methods, while also maintaining a competitive performance on te (mtg). 
In contrast, other augmentation methods (\synonym{} and \random) are harmful compared to \noaug{} on all treebanks, indicating that the examples generated by \sub{} may be closer to the original data distribution. 

\subsection{Dependency Parsing}
\begin{table}[t]
\centering \small
\begin{tabular}{lrrrrr}
\toprule 
$|{\mathcal{D}'}| \backslash |\mathcal{D}|$ & 10 & 50 & 100 & 500 & 1,000\\
\midrule 
\noaug & 26.1 & 53.1 & 63.1 & 79.9 & 83.1 \\
\midrule 
$2\times$ (\synonym) & 28.2 & 51.2 & 60.8 & 78.6 & 82.4 \\
$5\times$ (\synonym) & 25.8 & 50.3 & 61.3 & 77.2 & 80.8 \\
$10\times$ (\synonym) & 26.6 & 48.3 & 58.9 & 76.2 & 80.2 \\
$50\times$ (\synonym) & 27.0 & 49.6 & 59.4 & 77.0 & 79.1 \\
$100\times$ (\synonym) & \bf 28.9 & 44.8 & 59.4 & 72.9 & 76.8 \\
\midrule 
$2\times$ (\random) & 25.8 & 53.1 & 60.2 & 79.1 & 80.1 \\
$5\times$ (\random) & 28.6 & 54.0 & 61.5 & 78.4 & 82.5 \\
$10\times$ (\random) & 28.3 & 55.2 & 66.2 & 78.4 & 83.2 \\
$50\times$ (\random) & 28.1 & 54.4 & 58.4 & 78.5 & 83.1 \\
$100\times$ (\random) & 28.4 & 53.4 & 57.5 & 77.2 & 80.5 \\
\midrule 
$2\times$ (\sub) & 26.2 & 51.8& 63.6& 79.6 & 82.8 \\
$5\times$ (\sub) & 25.1 & 54.5 & 63.5 & 79.7 & 83.6 \\
$10\times$ (\sub) & 27.7 & 54.5 & 66.2 & 80.6 & 83.6 \\
$50\times$ (\sub) & 28.5 & 55.3 & \bf 68.5 & \bf 81.4 & \bf 84.2 \\
$100\times$ (\sub) & 28.1 & \bf 57.6 & 67.0 & 80.4 & 83.7 \\
\bottomrule 
\end{tabular}
\caption{\label{tab:d-parse-size} Labeled attachment scores (LAS) on the PTB test set. We start with an original training set $\mathcal{D}$, which consists of $|\mathcal{D}| \in \{10, 50, 100, 500, 1000\}$ examples, and augment it for $k \in \{2, 5, 10, 50, 100\}$ times. For each training set $\mathcal{D}$, the corresponding development set consists of $\max\left(10, \frac{|\mathcal{D}|}{10}\right)$ examples. }
\end{table} 

\begin{table}[t]
    \centering\small 
    \begin{tabular}{lrr}
    \toprule
         Model &  UAS & LAS \\
    \midrule
         \citet{mrini-etal-2020-rethinking}$^{\dagger}$ & 97.4 & 96.3 \\
    \midrule
         \citet{zhang-etal-2020-efficient}$^{\ddagger}$ & 96.1 & 94.5  \\
         BiAffine+XLM-R & 96.7 & 95.2 \\
         BiAffine+XLM-R+\sub & 96.6 & 95.2 \\ 
    \bottomrule
    \end{tabular}
    \caption{Unlabeled attachment score (UAS) and labeled attachment score (LAS) on the PTB dependency test set. Models are trained with the full PTB training set. $\dagger$: the previously best result using any kind of annotation (e.g., constituency parse trees); $\ddagger$: the previously best result using only dependency annotations. BiAffine: the bi-affine dependency parsing model proposed by \citet{dozat2016deep}.}
    \label{tab:d-parse-full}
\end{table}
We evaluate the performance of models using the standard Penn Treebank dataset \citep[PTB;][]{marcus-etal-1993-building}, converted by Stanford dependency converter v3.0,\footnote{\url{https://nlp.stanford.edu/software/stanford-dependencies.shtml}} following the standard splits. 

We first compare the performance of \sub{} and baselines in the low-resource setting (Table~\ref{tab:d-parse-size}). 
All methods, though not always, may help achieve better performance than \noaug.
\synonym{} helps achieve the best LAS when there is only an extremely small training set (e.g., 10 examples) available; however, when the size of the original training set becomes larger, \sub{} begins to dominate, while \synonym{} and \random{} start to sometimes hurt the performance. 
In addition, a larger augmented dataset does not necessarily lead to better performance -- throughout our experiments, augmenting the original dataset to $10\times$--$50\times$ larger can result in reasonably good accuracy. 

However, when training on the full WSJ training set, \sub{} does not necessarily help improve over baselines, but the performance is quite competitive (Table~\ref{tab:d-parse-full}).\footnote{An additional finding here is that a simple biaffine dependency parsing model \citep{dozat2016deep} with XLM-R initialization is able to set a new state of the art for dependency parsing with only in-domain annotation.}

\subsection{Constituency Parsing}
We evaluate \sub{} and baseline methods on few-shot constituency parsing, using the Foreebank \citep[Fbank;][]{kaljahi-etal-2015-foreebank} and NXT-Switchboard \citep[SWBD;][]{calhoun2010nxt} datasets. 
Foreebank consists of 1,000 English and 1,000 French sentences; for either language, we randomly select 50 sentences for training, 50 for development, and 250 for testing.\footnote{We leave the other 650 sentences for future use.} 
We follow the standard splits of NXT-Switchboard, and randomly select 50 sentences from the training set and 50 from the development set for training and development respectively. 

We compare different data augmentation methods using the setup of few-shot parsing from scratch (Table~\ref{tab:c-parse-few}).
Among all settings we tested, \sub{} achieves the best performance, while all augmentation methods we investigated improve over training only on the original dataset (\noaug). 
Surprisingly, we find that the seemingly meaningless \random{}, which random shuffles the sentence and inserts the shuffled words back into the original parse tree structure as the nonterminals, also consistently helps few-shot parsing by a nontrivial margin.\footnote{This trend may be explained by benefits in learning/optimization stability in this few-shot setting, but we leave a richer exploration of potential explanations for future work. 
}

For domain adaptation (Table~\ref{tab:c-parse-transfer}), we first train Benepar \citep{kitaev-klein-2018-constituency} on the Penn Treebank dataset, and use the pretrained model as the initialization. 
While compared to few-shot parsing trained from scratch, the gain by data augmentation generally becomes smaller, \sub{} still works the best across datasets. 

\begin{table}[t]
    \centering \small 
    \begin{tabular}{lrrr}
        \toprule
        Method & Fbank (en) & Fbank (fr) & SWBD \\
        \midrule
        \noaug & 33.1 & 27.3 & 29.1 \\
        \synonym & 64.8 & 59.9 & 51.1 \\
        \random & 55.9 & 48.8 & 37.0 \\
        \sub  & \bf 71.8 & \bf 70.8 & \bf 64.6\\
        \bottomrule
    \end{tabular}
    \caption{Labeled $F_1$ scores ($\times 100)$ on the test set of each constituency treebank, in the setting of few-shot parsing. The best number in each column is bolded. }
    \label{tab:c-parse-few}
\end{table}

\begin{table}[t]
    \centering \small 
    \begin{tabular}{lrrr}
        \toprule
        Method & Fbank (en) & Fbank (fr) & SWBD \\
        \midrule
        PTB  & 82.3 & 30.8 & 74.3 \\
        (PTB$\rightarrow\cdot$) \noaug &  83.1 & 70.1 & 77.2\\
        (PTB$\rightarrow\cdot$) \synonym  & 84.0 & 71.1 & 78.2\\
        (PTB$\rightarrow\cdot$) \random  & 83.5 & 70.1 & 75.6\\
        (PTB$\rightarrow\cdot$) \sub  & \bf 84.6 & \bf 72.6 & \bf 78.3\\
        \bottomrule
    \end{tabular}
    \caption{Labeled $F_1$ scores ($\times 100)$ on the test set of each constituency treebank, in the setting of domain adaptation. PTB: directly testing the model trained on the Penn Treebank. The best number in each column is bolded. }
    \label{tab:c-parse-transfer}
\end{table}

\subsection{Text Classification}
\begin{table}[t]
    \centering\small
    \begin{tabular}{lr}
        \toprule
        Method & Accuracy\\
        \midrule
        \multicolumn{2}{l}{SST-10\% ~~($|\mathcal{D}_\textit{train}|=0.8\text{K}, |\mathcal{D}_\textit{dev}|=0.1\text{K}$)}\\
        \midrule 
        \noaug & 25.4\\
        \synonym & 31.9 \\ 
        \random & 24.7 \\ 
        \randomword & 30.2 \\
        \sub+\textsc{p+t} (balanced tree) & 42.8 \\ 
        \sub+\textsc{p+t+senti} & \bf 46.0 \\
        \sub+\textsc{p} & 39.7 \\
        \sub+\textsc{p+l} & 43.1 \\
        \sub+\textsc{p+l+n} & 44.3 \\
        \sub+\textsc{p+l+n+t} & 43.4 \\
        \sub+\textsc{p+l+t} & 44.1 \\
        \sub+\textsc{p+t} & 45.0 \\
        \sub+\textsc{p+n+t} & 44.3 \\
        \sub+\textsc{p+n} & 43.7 \\
        \sub+\textsc{n+t} & 34.6\\
        \sub+\textsc{t} & 24.7 \\
        \midrule 
        \multicolumn{2}{l}{AG News-1\% ~~($|\mathcal{D}_\textit{train}|=0.6\text{K}, |\mathcal{D}_\textit{dev}|=0.06\text{K}$)}\\
        \midrule 
        \noaug & 40.6 \\
        \synonym & 86.2 \\ 
        \random & 80.8 \\ 
        \randomword & 82.1 \\
        \sub+\textsc{p} (balanced tree) & 85.9 \\ 
        \sub+\textsc{p} & 85.7 \\
        \sub+\textsc{p+l} & 85.4 \\
        \sub+\textsc{p+l+n} & 86.1 \\
        \sub+\textsc{p+l+n+t} & 86.1 \\
        \sub+\textsc{p+l+t} & 86.7 \\
        \sub+\textsc{p+t} & \bf 86.8 \\
        \sub+\textsc{p+n+t} & \bf 86.8\\
        \sub+\textsc{p+n} & 85.6 \\
        \sub+\textsc{n+t} & 86.4 \\
        \sub+\textsc{t} & 82.4 \\
        \bottomrule
    \end{tabular}
    \caption{Accuracy ($\times 100$) on the AG News sentence test set and SST standard test set. The best numbers in each section are bolded. } 
    \label{tab:txt-class}
\end{table}

\begin{table}[t]
    \centering \small
    \begin{tabular}{lrr}
    \toprule
    Method & Dev. Acc. & Test Acc.  \\
    \midrule
    XLM-R (\noaug) & 56.1 & 55.7 \\
    XLM-R (\sub) & \bf 56.6 & \bf 56.6 \\
	\midrule
	\citet{brahma2018improved} & N/A & 56.2 \\
    \bottomrule
    \end{tabular}
    \caption{Accuracy ($\times 100$) on the SST standard development and test set. }
    \label{tab:sst_full}
\end{table}

We evaluate the methods introduced in Section~\ref{sec:var} and baselines on two text classification datasets: \citep[SST;][]{socher-etal-2013-recursive} and AG News \citep{zhang2015character} sentence (Table~\ref{tab:txt-class}), in the low-resource setting.\footnote{We only keep the single-sentence instances among all examples in each split of the original AG News dataset, following \citet{shi-etal-2018-tree}. } 
We obtain the constituency parse trees using Benepar \citep{kitaev-klein-2018-constituency} trained on the standard PTB dataset. 
Since the SST dataset provides sentiment labels of phrases, it is also natural to apply such phrase sentiment labels as substructure labels, where the substructures are phrases (\sub+\textsc{p}+\textsc{senti}). 

Across the two investigated settings, data augmentation is usually helpful to improve over \noaug, and most variations of \sub{} with the phrase-or-not (+\textsc{p}) substructure label are among the best-performing methods on each task (except \sub+\textsc{p} for SST-10\%). 
Additionally, constituency tree--based \sub{} with phrase labels (+\textsc{p+l}) outperforms balanced tree--based \sub{} in both settings, indicating that phrase structures can be considered as useful information for data augmentation in general. 

We further use \sub+\textsc{p+t+senti} to augment the full SST training set, since it is the best augmentation method for few-shot sentiment classification. 
In addition to sentences, we also add phrases (i.e., subtrees) as training examples, following most of existing work \interalia{socher-etal-2013-recursive,kim-2014-convolutional,brahma2018improved},\footnote{That is, different from Table~\ref{tab:txt-class}, we apply the same settings as conventional work to produce numbers in Table~\ref{tab:sst_full}. } to boost performance. 
In this setting, we find that \sub{} helps set a new state of the art on the SST dataset (Table~\ref{tab:sst_full}).

\section{Discussion}
We investigate substructure substitution (\sub), a family of data augmentation methods that generates new examples by same-label substructure substitution. 
Such methods help achieve competitive or better performance on the tasks of part-of-speech tagging, few-shot dependency parsing, few-shot constituency parsing, and text classification.
While other data augmentation methods (e.g., \synonym{} and \random) sometimes improve the performance, \sub{} is the only one that consistently helps low-resource NLP across tasks. 


While existing work has shown that explicit constituency parse trees may not necessarily help improve recursive neural networks for text classification and other NLP tasks \citep{shi-etal-2018-tree}, our work shows that such parse trees can be robust backbones for \sub-style data augmentation, introducing more potential ways to help neural networks take advantages from explicit syntactic annotations. 

There is an open question remaining to be addressed: it is still unclear that why \random{} helps improve few-shot constituency parsing, as the training process requires the model to output the correct parse tree of a sentence while only accessing shuffled words. 
We leave the above question, as well as applications of \sub{} to more NLP tasks, for future work. 


\bibliography{custom}
\bibliographystyle{acl_natbib}

\end{document}